\documentclass[10pt,twocolumn,letterpaper]{article}

\usepackage{iccv}              
\usepackage{cuted}

\definecolor{iccvblue}{rgb}{0.21,0.49,0.74}
\usepackage[pagebackref,breaklinks,colorlinks,allcolors=iccvblue]{hyperref}
\usepackage{booktabs}  
\usepackage{pifont}    
\usepackage{caption}   
\usepackage{array}     
\usepackage{multirow}  
\usepackage{colortbl}
\usepackage{xcolor}
\definecolor{cvprblue}{RGB}{0, 82, 147}
\usepackage{arydshln}
\usepackage{graphicx}
\usepackage{lipsum}
\usepackage{indentfirst}
\usepackage{amsmath}
\usepackage{cases}
\usepackage{booktabs}
\usepackage{multirow} 
\usepackage{graphicx}
\usepackage{times}  
\usepackage{algorithm}
\usepackage{algorithmic}
\usepackage{color}
\usepackage{ulem}
\usepackage{pifont}
\usepackage{arydshln} 
\usepackage{float} 
\usepackage{bm}
\usepackage{caption}
\usepackage{listings}

\def\paperID{13032} 
\def\confName{ICCV}
\def\confYear{2025}

\title{LayerTracer: Cognitive-Aligned Layered SVG Synthesis via  Diffusion Transformer}

\author{
Yiren Song \quad Danze Chen \quad Mike Zheng Shou\thanks{Corresponding author.} \\
Show Lab, National University of Singapore \\
}

\begin{document}
\maketitle


\begin{strip}
\centering
    \vspace{-16mm}
    \includegraphics[width=\linewidth]{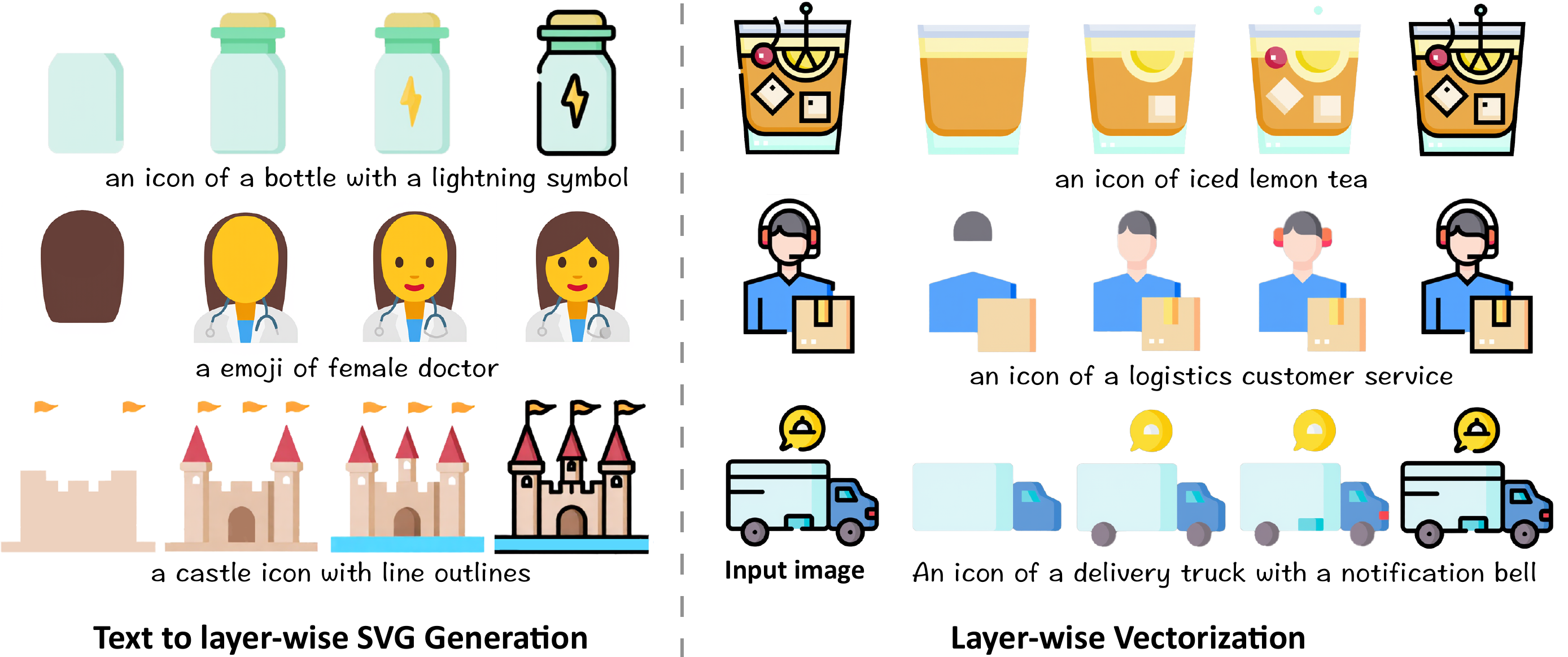}
    \vspace{-6mm}
    \captionof{figure}{LayerTracer creates cognitively aligned layered SVGs from text prompts or by converting images into layered SVGs.}
    \label{teaser}
\end{strip}









\begin{abstract}
\vspace{-6mm}

Generating cognitive-aligned layered SVGs remains challenging due to existing methods’ tendencies toward either oversimplified single-layer outputs or optimization-induced shape redundancies. We propose LayerTracer, a DiT based framework that bridges this gap by learning designers’ layered SVG creation processes from a novel dataset of sequential design operations. Our approach operates in two phases: First, a text-conditioned DiT generates multi-phase rasterized construction blueprints that simulate human design workflows. Second, layer-wise vectorization with path deduplication produces clean, editable SVGs. For image vectorization, we introduce a conditional diffusion mechanism that encodes reference images into latent tokens, guiding hierarchical reconstruction while preserving structural integrity. Extensive experiments show that LayerTracer surpasses optimization-based and neural baselines in generation quality and editability. Code is released at \href{https://github.com/showlab/LayerTracer}{https://github.com/showlab/LayerTracer}.

\end{abstract}

\section{Introduction}


Scalable Vector Graphics (SVG) is widely used in modern digital design, representing visual elements such as paths, curves, and geometric shapes through mathematical equations rather than pixel grids. Unlike raster images, SVG preserves resolution-independent clarity at any scale, making it suitable for applications requiring high precision, including UI/UX design and industrial CAD systems. Layered SVGs further enhance this flexibility by allowing designers to manipulate individual layers to adjust stroke properties, spatial arrangements, and compositing effects. This structured editability supports dynamic modifications and collaborative workflows in contemporary design practices.

Nevertheless, a significant gap persists between current deep learning-based SVG generation techniques and professional requirements. Existing approaches face three systemic challenges: First, the scarcity of large-scale layered SVG datasets forces models to rely on synthetic or oversimplified training data, resulting in outputs devoid of the nuanced hierarchical structures inherent to human designs. Second, methodological fragmentation prevails—optimization-based methods \cite{vectorfusion, svgdreamer, diffsketcher, live, diffvg, oar, layered_vectorization} generate vector paths using raster priors but often produce cluttered geometries with redundant anchor points; and large language models (LLMs) \cite{empoweringLLM, strokenuwa, starvector,  xing2024llm4svg}, constrained by token limits, remain limited to basic icons . Most critically, no existing method addresses the designer’s cognitive process—the logical sequencing, spatial reasoning, and element grouping strategies employed during layer construction—resulting in AI-generated SVGs that resemble fragmented collages rather than intentionally editable professional designs.

To address these challenges, we present LayerTracer, a Diffusion Transformer (DiT)-based framework that redefines layered SVG synthesis by modeling designers’ layer-by-layer construction logic. Our approach is grounded in three key insights: (1). Cognitive alignment: DiT models pretrained on text-image corpora inherently capture contextual relationships between visual elements, which can be steered through targeted fine-tuning to mimic designer decision-making. (2). Spatiotemporal consistency: The self-attention mechanism’s bias toward local token interactions—a byproduct of training on natural image pixel correlations—can be repurposed to enforce coherence across sequential design steps. 3. Structured decomposition: Disassembling layered SVGs into channel-wise components and organizing them as grid sequences provides generation models with an interpretable blueprint of layer evolution.

In implementation, LayerTracer integrates two innovations. First, we curate a pioneering dataset of 20,000+ designer process traces, automatically converting layered SVGs into timestamped creation sequences. These sequences are rasterized and organized into training grids using a serpentin layout, ensuring temporally adjacent design steps remain spatially proximate. Second, we develop a dual-phase generation pipeline: (1) a text-conditioned DiT generates rasterized construction process sequences that simulate a designer’s workflow, followed by (2) a layer-wise vectorization module that converts these sequences into clean, editable SVG layers while eliminating redundant paths.

Beyond text-to-SVG synthesis, LayerTracer tackles the inverse task: converting raster images into layered vector graphics. We reframe this as a process-conditioned generation problem, where reference images guide the model to "reverse-engineer" plausible layer construction steps. Specifically, we build upon a pretrained DiT model and adapt it through LoRA fine-tuning to ingest image context. By encoding reference images into conditional tokens injected into the denoising process, the model autonomously deduces layer assembly sequences (e.g., "background first, then foreground elements"), faithfully reconstructing input images while adhering to practical editing constraints.

Our main contributions are as follows:
\begin{itemize}
\item Cognitive-aligned SVG synthesis: As the first framework to generate layered SVGs by learning designers’ construction logic—element ordering, layer grouping, and spatial reasoning—LayerTracer ensures outputs meet professional editing standards.
\item Unified DiT-based architecture: Our framework seamlessly integrates text-to-SVG generation and layer-wise vectorization tasks, eliminating the need for task-specific pipelines.
\item Process-centric dataset: We release a scalable pipeline for collecting designer workflow data, addressing the critical gap in layered vector graphics training resources. Extensive experiments validate LayerTracer’s state-of-the-art performance and effectiveness.
\end{itemize}









\section{Related Works}

\subsection{Text2image Diffusion Model}

Recent studies have demonstrated that diffusion models are capable of generating high-quality synthetic images, effectively balancing diversity and fidelity. Models based on diffusion models or their variants, such as those paper in  \cite{sd, dit}, have successfully addressed the challenges associated with text-conditioned image synthesis. Stable Diffusion  \cite{sd}, a model based on the Latent Diffusion Model, incorporates text conditioning within a UNet framework to facilitate text-based image generation \cite{diffsim, idp, anti}, establishing itself as a mainstream model in image generation. Fine-tuning pre-trained image generation models can enhance their adaptation to specific application scenarios, as seen in techniques like LoRA  \cite{lora} and DreamBooth  \cite{dreamfusion}. For theme control in text-to-image generation, several works  \cite{ipa, instantid, ssr, fast_icassp, makeup, hair, chen2025transanimate} focus on custom generation for defined pictorial concepts, with ControlNet  \cite{controlnet} additionally offering control over other modalities such as depth information. Some methods \cite{wang2025diffdecompose, zhang2025easycontrol, huang2025photodoodle, song2025omniconsistency, guo2025any2anytryon, shi2024fonts, lu2025easytext, gong2025relationadapter, shi2025wordcon, song2025makeanything} explore controllable generation using Diffusion Transformer-based approaches. AnimateDiff \cite{guo2023animatediff} introduces a temporal attention module, extending Stable Diffusion into a video generation model. Inspired by ProcessPainter \cite{processpainter}, which first proposed learning an artist's painting process through pre-trained temporal models, this paper leverages the in-context capabilities of DiT to generate a layered SVG creation process.


\subsection{SVG Generation }
Scalable Vector Graphics (SVGs) are widely utilized in design owing to their advantages like geometric manipulability, resolution independence, and compact file structure. SVG generation often involves training neural networks to produce predefined SVG commands and attributes using architectures such as RNNs \cite{im2vec}, VAEs \cite{svg2, svg+vae, strokenuwa}, and Transformers \cite{svg2, deepvecfont, iconshop}. Nonetheless, the absence of large-scale vector datasets constrains their generalization capabilities and the creation of complex graphics, with most datasets focusing on specific areas like monochromatic vector icons \cite{iconshop} and fonts \cite{deepvecfont, clipfont}.

An alternative to directly training an SVG generation network is optimizing it to match a target image during the evaluation phase, employing differentiable rasterizers to bridge vector graphics and raster images \cite{diffvg}. This method optimizes SVG parameters based on pretrained vision-language models. Advances in models like CLIP \cite{clip} have facilitated effective SVG generation methods such as CLIPDraw \cite{clipdraw}, CLIPasso \cite{clipasso}, and CLIPVG \cite{clipvg}. while DreamFusion \cite{dreamfusion} demonstrates the superior generative capabilities of diffusion models. VecFusion \cite{vectorfusion}, DiffSketcher \cite{diffsketcher}, and SVGDreamer combine differentiable rasterizers with text-to-image diffusion models to produce vector graphics, achieving notable results in iconography and sketching. However, these methods still face challenges with editability and graphical quality. Recent studies \cite{nivel, t2v} have blended optimization-based methods with neural networks to enhance vector representations by integrating geometric constraints. 

The primary issue with methods that optimize a set of vector primitives through SDS loss is their reliance on image generation model priors, which often leads to redundant and noisy results. These outputs lack clear hierarchical structures and fail to meet design specifications. In this paper, we innovatively propose an alternative approach to utilizing image generation model priors. Specifically, we leverage the in-context learning capability of Diffusion Transformers to generate the creation process of SVG graphics, combined with vectorization to achieve cognitive-aligned layered SVG generation.




\subsection{Vectorization}
Raster image vectorization or image tracing is a well-studied problem in computer graphics\cite{svg1, svg2, svg3, svg4}. Diffvg\cite{diffvg} proposes a differentiable rendering method for vectorization, which found shape gradients by differentiating the formula of Reynolds transport theorem with Monta-Carlo edge sampling. Meanwhile, combining differentiable rendering techniques with deep learning models are also studied for image vectorization\cite{live, clipasso, cliptexture}. Direct raster-to-vector conversion with neural networks are supported for the relatively simple images\cite{svg+vae, svg2, im2vec}.  Stroke-based rendering can be used to fit a complex image with a sequence of vector strokes \cite{singh2022intelli, liu2021paint, hu2023stroke}, but the performance is limited by the predefined strokes. Diffvg\cite{diffvg} can also be leveraged to fit an input image with a set of randomly initialized vector graphical elements. Based on Diffvg, LIVE \cite{live} proposes a coarse-to-fine vectorization strategy, with cost tens of minute. CLIPVG\cite{clipvg} proposes a multi-round vectorization strategy,  providing additional graphic elements for the image manipulation task. LIVE \cite{live} and O\&R \cite{oar} achieve hierarchical vectorization through optimization-based methods, but their results show a significant gap compared to human-designed works, lacking logical coherence. In contrast to these approaches, our proposed LayerTracer leverages the prior knowledge of the Diffusion Transformer model, reformulating the hierarchical vectorization task as a problem of predicting preceding frames from a reference image.


\begin{figure*}[ht]
    \centering
    \includegraphics[width=1\linewidth]{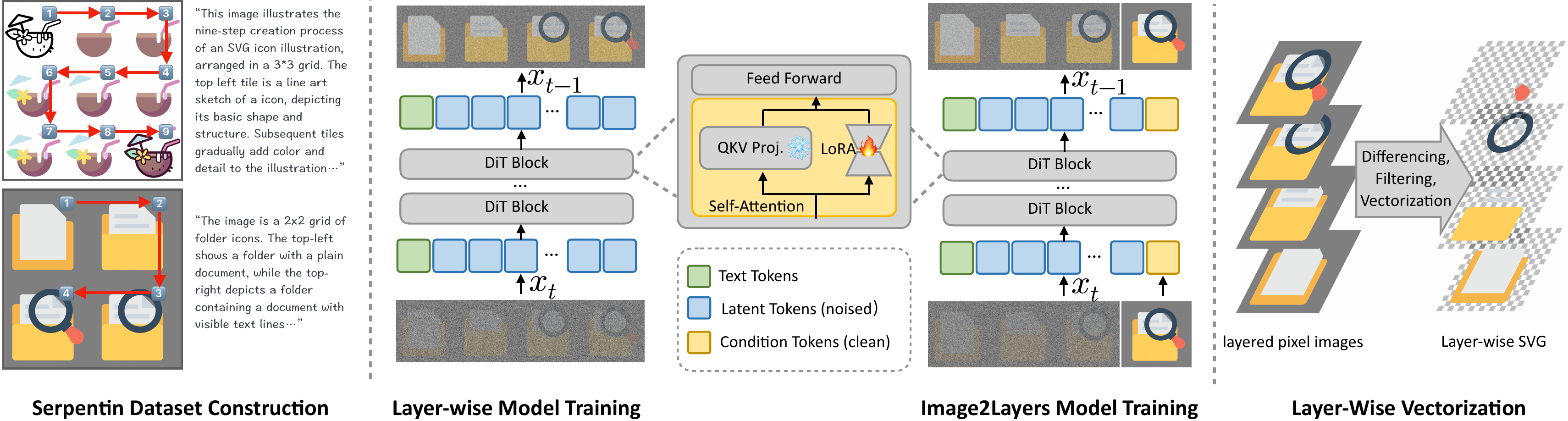} 
    \caption{The LayerTracer architecture comprises three key components: (1) \textbf{Layer-wise Model}: Pretrained on our proposed dataset to generate layered pixel sequences from text prompt; (2) \textbf{Image2Layers Model}: Merges LoRA with the Flux base DiT, enabling image-conditioned generation through VAE-encoded latent tokens; (3) \textbf{Layer-wise Vectorization}: Converts raster sequences to SVGs via differential analysis between adjacent layers, followed by Bézier optimization using vtracer to eliminate redundant paths while preserving structural fidelity.}
    \label{method}
\end{figure*}

\section{Method}

In this section, we begin by exploring the preliminaries on diffusion transformer as
detailed in section 3.1. Then introduce the overall architecture of our method in section 3.2, followed by detailed descriptions of the key modules: dataset construction methods in 3.3, Layer-wise image generation in section 3.4, Image2Layers in section 3.5, and Layer-Wise Vectorization in section 3.6. 






\subsection{Preliminary}

The Diffusion Transformer (DiT) model \cite{dit}, which appears in frameworks such as FLUX.1 \cite{flux2023}, Stable Diffusion 3 \cite{sd}, and PixArt~\cite{pixart}, employs a transformer-based denoising network to iteratively refine noisy image tokens.

DiT processes two categories of tokens: noisy image tokens $X \in \mathbb{R}^{N \times d}$ and text condition tokens $C_T \in \mathbb{R}^{M \times d}$, where $d$ is the embedding dimension, and $N$ and $M$ respectively represent the numbers of image and text tokens. As these tokens move through the transformer blocks, they retain consistent dimensions.

In FLUX.1, each DiT block applies layer normalization before Multi-Modal Attention (MMA) \cite{mma}, incorporating Rotary Position Embedding (RoPE) \cite{rope} to capture spatial context. For image tokens $X$, RoPE applies rotation matrices based on a token’s position $(i,j)$ in the 2D grid:
\begin{equation}
X_{i,j} \rightarrow X_{i,j} \cdot R(i,j),
\end{equation}
where $R(i,j)$ is the rotation matrix at position $(i,j)$. Text tokens $C_T$ are similarly transformed with their positions specified as $(0,0)$.

The multi-modal attention mechanism then projects these position-encoded tokens into query $Q$, key $K$, and value $V$ representations, enabling attention across all tokens:
\begin{equation}
\text{MMA}([X; C_T]) = \text{softmax}\left(\frac{QK^\top}{\sqrt{d}}\right)V,
\end{equation}
where $[X; C_T]$ denotes concatenation of image and text tokens. This formulation ensures bidirectional attention among the tokens.

\subsection{Overall Architecture.}


LayerTracer consists of the following components: Serpentine dataset construction, Layer-wise model training, Image2Layers model, and the layer-wise vectorization. Initially, we collected the processes designers use to create layered vector graphics and organized them into grid-based datasets. Following this, we utilized the LoRA method for pre-training on the proposed dataset, thereby enabling the generation of layered pixel images from textual descriptions. Subsequently, we integrated the LoRA from the previous step with the Flux base model to establish a new foundational model. The Image2Layers model introduces an image-based conditional mechanism that, through additional LoRA fine-tuning, predicts the creation process of reference images. Finally, in the layer-wise vectorization stage, the model sequentially transforms the generated pixel images into high-quality vector graphics, which are analyzed, filtered, and vectorized based on the differences between adjacent layers.

\subsection{Serpentine Dataset Construction}
Our dataset construction includes 20,000 layered SVGs created by designers, encompassing black outline icons, regular icons, emojis, and illustrative graphics. As shown in Fig. \ref{method}, each sequence is composed of either 9 or 4 frames, arranged in 3x3 or 2x2 grids, resulting in resolutions of 1056x1056 and 1024x1024 respectively. To capture the process of designers creating layered SVGs, we propose a automated data generation pipeline that deconstructs the layered SVG graphics into sequences based on the grouping logic and element hierarchy within the SVG files. Additionally, the pipeline incorporates a human-in-the-loop process to filter out nonsensical sequences.

In the attention mechanism of DiT, tokens tend to focus on spatially adjacent tokens. This tendency stems from the strong correlations between adjacent image pixels captured during the pre-training of diffusion models. To enhance the model's learning of grid sequences \cite{grid}, we introduce the serpentine dataset construction. As shown in Figure \ref{method}, we arrange the sequences of 9 and 4 frames in a serpentine layout within the grid, ensuring that temporally adjacent frames are also spatially adjacent (either horizontally or vertically). In our ablation experiments, we confirmed that this design is crucial for the coherence of sequence generation. To aid hierarchical vectorization, icons with black strokes isolate the line layer in the first frame during dataset creation. During the generation phase, we vectorize the black line layer and overlay it onto the subsequent results.




\subsection{Layer-wise Image Generation} 

DiT models, trained on massive image-text pairs, inherently possess contextual generation capabilities. By appropriately activating and enhancing this ability, they can be utilized for complex generation tasks. Since text-to-image models can interpret merged prompts, they can be reused for in-context generation without altering their architecture. This only requires changes to the input data rather than modifications to the model itself. Building on this insight, we designed a simple yet effective pipeline to learn the hierarchical logic employed by human designers in creating layered SVGs.




\noindent \textbf{Layer-wise Model Training.} Due to the size of the dataset, we adopt LoRA fine-tuning for training which can be formulated as:
\begin{equation}
W = W_0 + \Delta W,
\end{equation}
where \(W_0\) represents the original weights of the pre-trained model, and \(\Delta W\) denotes the low-rank adaptation updates introduced during fine-tuning. This formulation enables efficient training by keeping \(W_0\) fixed and applying lightweight updates through \(\Delta W\), which allows the model to balance generalization from the pre-trained weights with task-specific adaptation provided by the fine-tuned updates.

\noindent \textbf{Loss function.} We employ the conditional flow matching loss function, integral to training and optimizing the generative model, is defined as follows:
{\setlength{\abovedisplayskip}{2pt}   
 \setlength{\belowdisplayskip}{2pt}   
 \begin{equation}
   L_{\mathrm{CFM}}
     = \mathbb{E}_{t,\;p_t(X|\epsilon),\;p(\epsilon)}
       \!\bigl[
         \bigl\|\,v_\Theta(X,t) - u_t(X|\epsilon)\bigr\|^2
       \bigr]
 \end{equation}
}
Where \( v_\Theta(X, t) \) represents the velocity field parameterized by neural network weights, \(t\) denotes the timestep, and \( u_t(X|\epsilon) \) is the conditional vector field mapping the path between noise and true data distributions.




\subsection{Image2Layers Model}

In this section, we introduce the Image2Layers model, which builds upon the previous section by incorporating image conditioning. This approach redefines hierarchical vectorization as a "reverse engineering" task, predicting how an SVG is created layer by layer.

The primary challenge in training the Image2Layers model lies in the limited availability of high-quality sequential data. While a small dataset may suffice for LoRA training, initializing a controllable plugin (such as ControlNet \cite{controlnet} or IP-Adapter \cite{ipa}) from scratch with limited data is highly challenging. To address this, we design an efficient controllability framework by repurposing a pre-trained DiT model and adapting it to accept image context as a conditioning input. Specifically:

\noindent \textbf{Training Phase.} We concatenate procedural sequences into 2×2 or 3×3 training grids. The final frame (context image) is passed through the VAE to extract latent variables, which are directly appended to the end of the denoising latent. Through self-attention mechanisms, the context latent provides conditional information to the denoising processes of other frames, enhancing the logical consistency and coherence of the generation. The condition image is then fed into a VAE to obtain its latent representation, which is directly appended to the denoising latent at the end. Multi-modal attention mechanisms are used to provide conditional information for the denoising of other frames. 
{\setlength{\abovedisplayskip}{4pt}  
 \setlength{\belowdisplayskip}{4pt}  
 \begin{equation}
   \text{MMA}([X; C_I; C_T]) = \text{softmax}\!\left(\frac{QK^\top}{\sqrt{d}}\right)V ,
 \end{equation}
}
where $[X; C_I; C_T]$ denotes the concatenation of image and text tokens. This formulation enables bidirectional attention.

\noindent  \textbf{Inference Phase.} During inference, we use the reference image as a condition to predict the earlier layers, thereby inferring how the SVG in the reference image was constructed layer by layer.


\begin{figure*}[ht]
    \centering
    \includegraphics[width=1.0\linewidth]{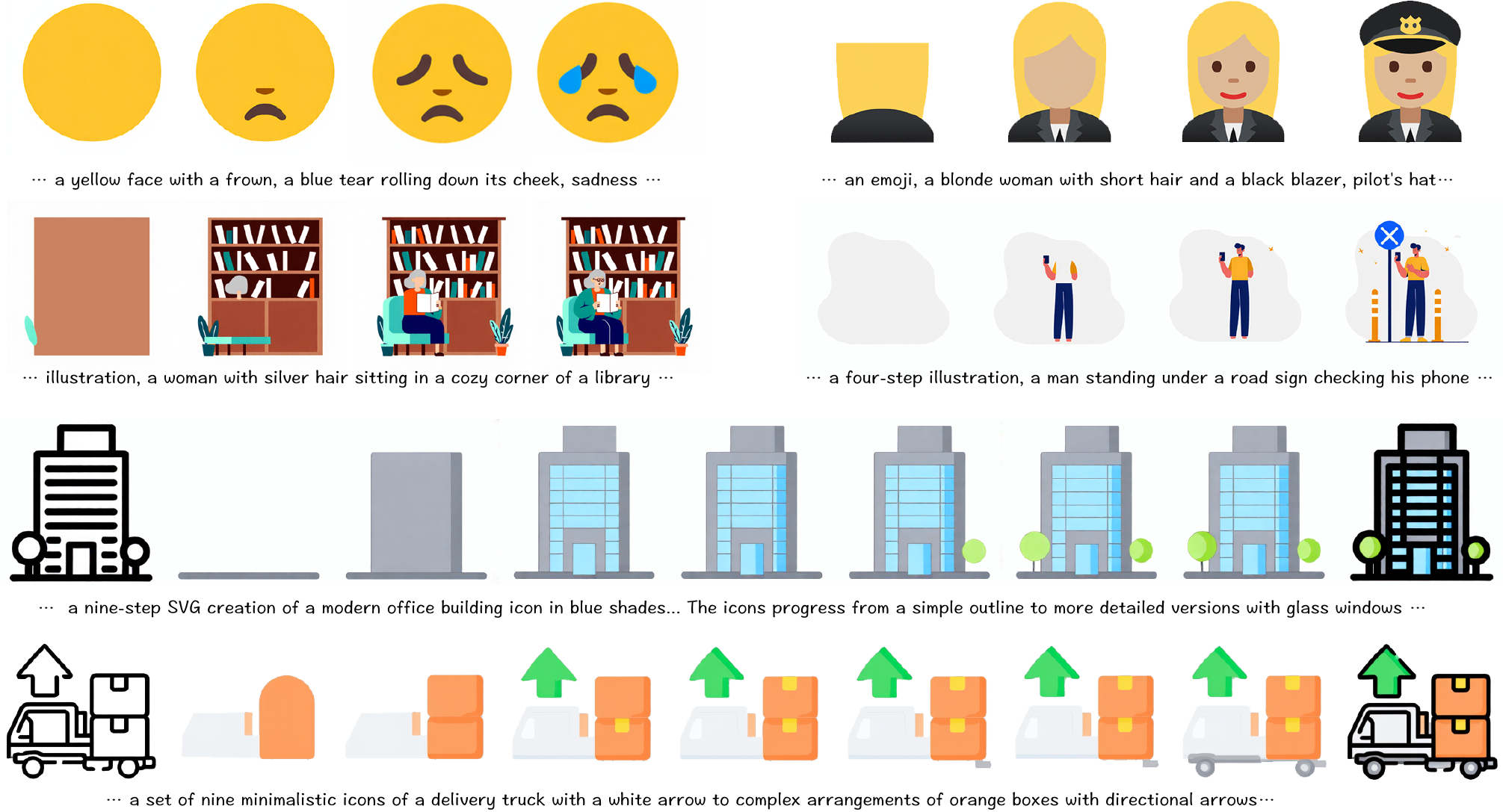} 
    \caption{Given a text prompt, LayerTracer generates cognitive-aligned layered SVGs that mimic human design cognition.}
    \label{fig3}
\end{figure*}


\subsection{Layer-Wise Vectorization}
To achieve hierarchical vectorization of input images, our process begins by segmenting grid images into individual cells. This facilitates independent processing for subsequent vectorization stages. Specifically for icons, accurate extraction of black lines is crucial as they represent key structural elements of the image. To address common issues of line distortion, we employ a series of preprocessing steps: grayscale conversion highlights contrast between lines and background; Gaussian blurring smooths out noise; and adaptive thresholding via the Otsu's thresholding method ensures robust line separation. These preprocessed lines are then integrated into a transparent PNG, focusing vectorization efforts on relevant areas.

Furthermore, to capture the layered details of images, we perform differential extraction between adjacent cells, identifying significant pixel changes to highlight areas of variation. This involves converting cell images to grayscale, computing absolute differences, applying binary thresholding, and refining the output with morphological operations to produce clean, meaningful contours of change. These contours are saved as transparent PNGs for subsequent vectorization.

The final step involves vectorizing these differential layers using tools like vtracer, optimizing parameters to balance detail retention and file size, and ultimately merging all vectorized layers into a single SVG file. This method preserves the image's global structure while highlighting intricate changes between cells, resulting in a layered and editable SVG suitable for detailed graphical representations.

\begin{figure*}[ht]
    \centering
    \includegraphics[width=1.0\linewidth]{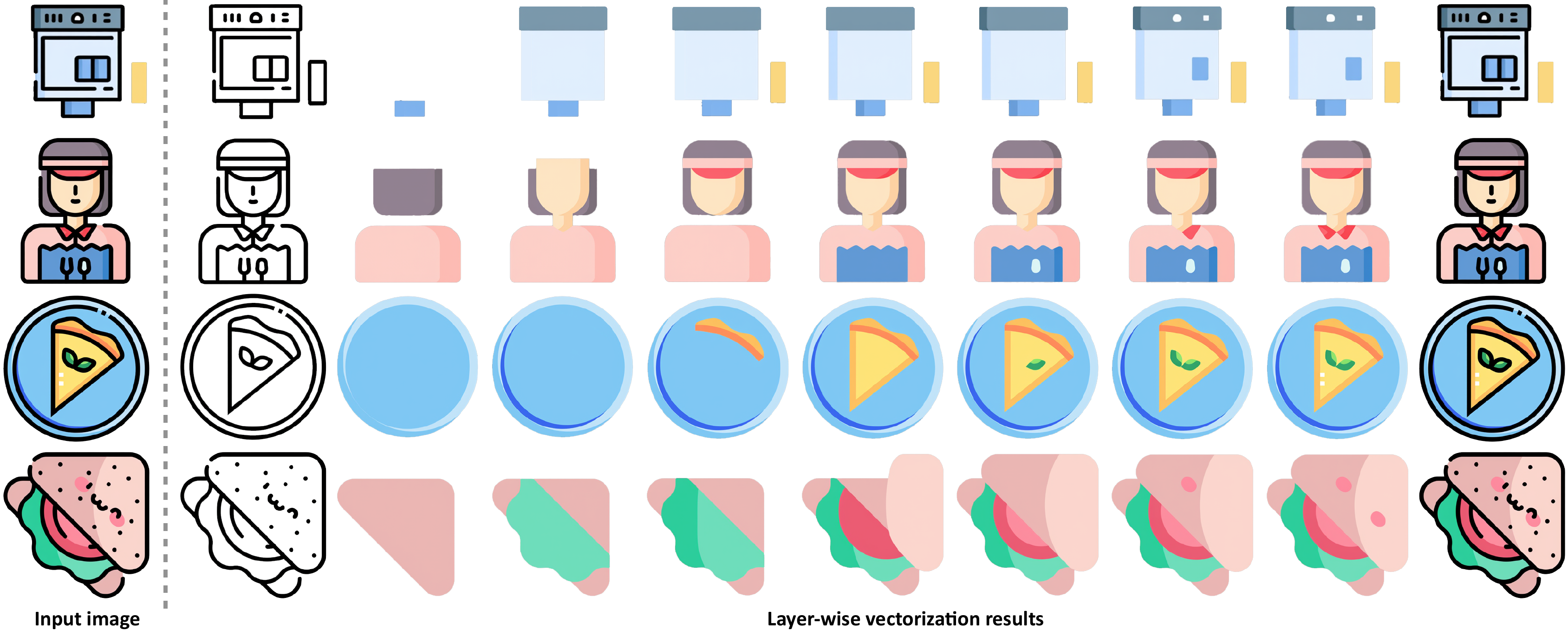} 
    \caption{Given a raster image of an icon as input, LayerTracer predicts how the icon was created layer by layer, achieving cognitive-aligned layered vectorization.}
    \label{fig4}
\end{figure*}

\section{Experiment}

\subsection{Experiment Setting}


\textbf{Experiment Details.}  
During the pretraining stage, we utilized the Flux 1.0 dev model based on the pretrained DiT architecture. Training resolutions included 1056×1056 (3×3 grids) and 1024×1024 (2×2 grids). The LoRA fine-tuning approach was applied with a LoRA rank of 256, a batch size of 16, a learning rate of 0.001, and 20,000 fine-tuning steps. For the three styles of SVGs, we trained separate LoRA models for black outline icons, illustrations, and emojis, using 3K, 2K, and 15K training samples, respectively. During the Image2Layers Model training phase, we merged the LayerTracer LoRA with the base model, using a LoRA merge weight of 1.0. Then, we fine-tuned it for 20,000 steps using LoRA-based training on the same dataset as the Layer-wise Model training. All training was conducted on a single A100 GPU with 80GB of memory.


\noindent \textbf{Baseline Methods.}  
The baseline methods in text-to-svg genration methods are SVGDreamer \cite{svgdreamer}, Vecfusion \cite{vectorfusion}and DiffSketcher \cite{diffsketcher}. The baseline methods in vectorization include diffvg \cite{diffvg}, LIVE \cite{live}, and O\&R \cite{oar}. All baseline methods are evaluated with their default settings to ensure fairness.


\noindent  \textbf{Benchmarks.}  To address the lack of high-quality layered SVG datasets, this paper introduces a dataset containing over 20,000 layered SVGs and their creation processes, named the LayerSVG Dataset. To ensure fairness in comparative experiments, the Noto-Emoji \cite{noto_emoji} dataset is also included in the benchmark for quantitative evaluation. For the text-to-SVG task and the layer-wise vectorization task, we select 50 prompts and 50 images, respectively, as benchmarks for testing.



\subsection{Generation Results}


\begin{figure}[!t]
    \centering
    \includegraphics[width=0.9\linewidth]{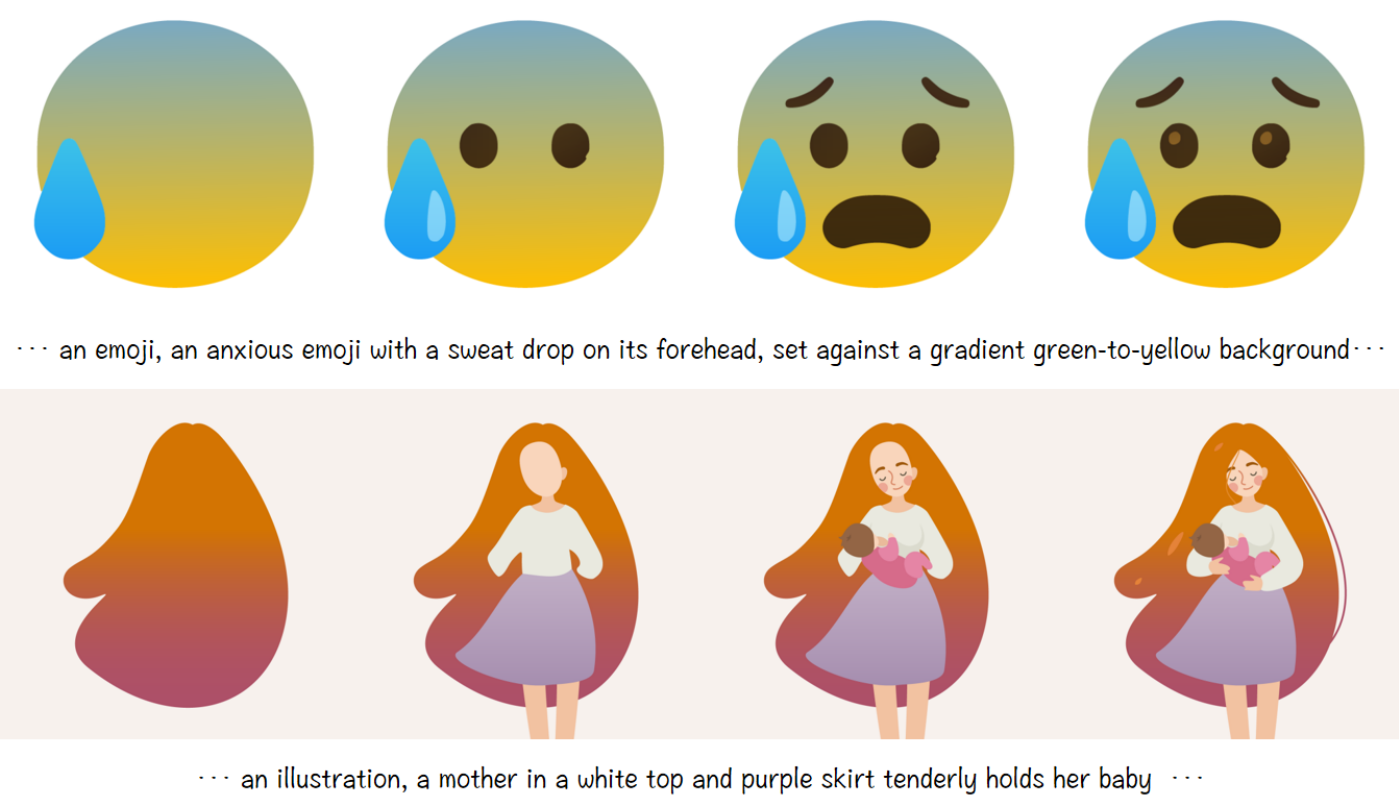}
    \caption{Layer-wise SVG generation with color gradients.}
    \label{gradients}
\end{figure}

Fig. \ref{fig3} demonstrates LayerTracer's capability to generate cognitively-aligned layered SVGs that adhere to text descriptions while maintaining logical layer hierarchies (e.g., background-to-foreground ordering and grouped semantic elements). The outputs preserve essential design properties including layer independence, non-overlapping paths, and topological editability. Fig. \ref{fig4} further illustrates layer-aware vectorization results, where input raster images are decomposed into clean vector layers with consistent spatial alignment and minimal shape redundancy. Fig. \ref{gradients} shows that training with gradient-colored samples and advanced vectorization methods \cite{du2023image} enables LayerTracer to generate layered vector graphics with color gradients. More experimental results can be found in the supplementary material.

\subsection{Comparison and Evaluation}
In the text-to-SVG task, we compute FID and CLIP Score. For the hierarchical vectorization task, we follow the evaluation methodology from previous works. We calculate MSE to assess the consistency between the reconstructed image and the input image. Additionally, we record the number of SVG shapes used, as fewer shapes indicate a more concise and efficient result. Table \ref{tab1} and \ref{tab2} show that LayerTracer achieves the best results across most metrics.



\begin{figure}[ht]
    \centering
    \includegraphics[width=1.0\linewidth]{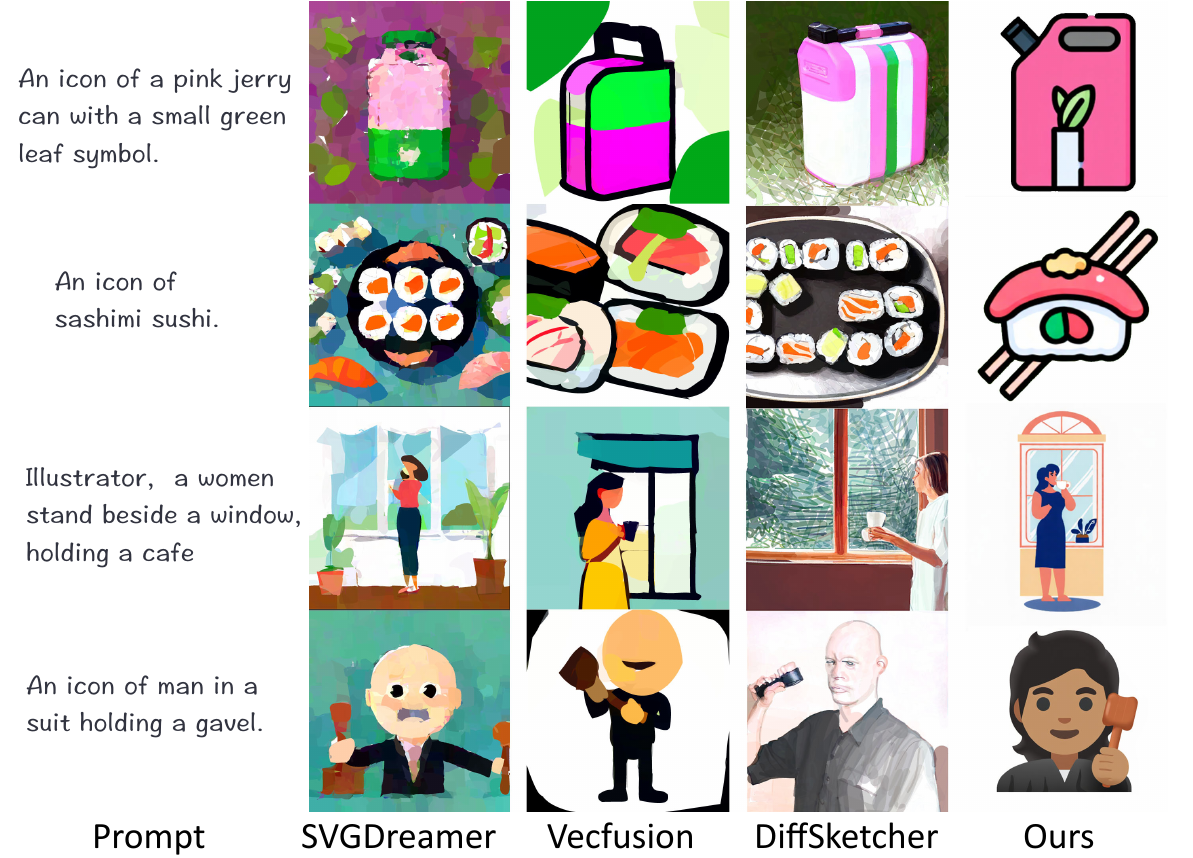} 
    \caption{Compare with baseline methods in Text-to-SVG generation task.}
    \label{fig5}
\end{figure}

\begin{figure*}[ht]
    \centering
    \includegraphics[width=1.0\linewidth]{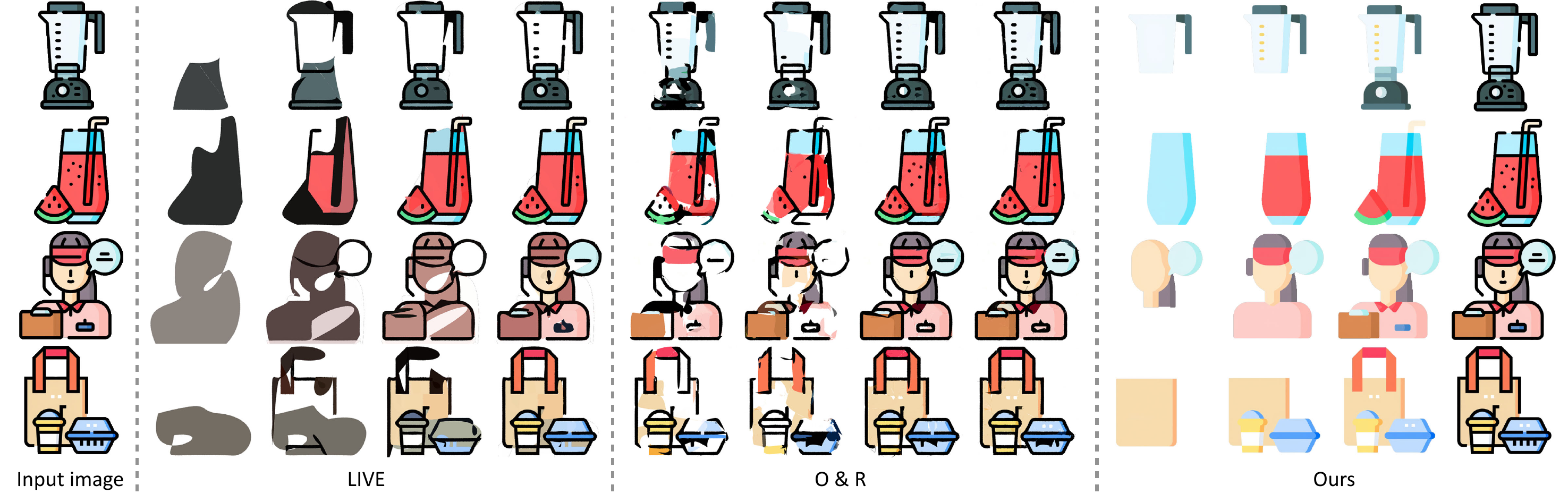} 
    \caption{Compare with baseline methods in layer-wise vectorization task. Our results are more concise and exhibit more logical layering.}
    \label{fig6}
\end{figure*}


\noindent \textbf{Qualitative Evaluation.} This section presents qualitative analysis results. Figure~\ref{fig5} shows that our method produces concise and coherent outputs for text-to-SVG tasks, aligning with specific requirements for icons and emojis. Baseline methods, in contrast, result in visual clutter and irregular paths. Figure~\ref{fig6} demonstrates our method's superior performance in layer-wise vectorization tasks, maintaining logical spatial hierarchies and semantic grouping, unlike baselines which yield fragmented and misaligned outputs.

\begin{table}[ht]
\centering
\footnotesize 
\caption{Comparison with SOTA SVG Generation Methods. The best results are denoted as \textbf{Bold}.}
\vspace{-3mm} 
\setlength{\tabcolsep}{4pt}
\begin{tabular}{lp{2.15cm}p{1cm}p{1cm}p{1cm}p{1cm}}

\toprule
& \textbf{Methods}   & \textbf{CLIP-Score}↑ & \textbf{Time Cost(s)}↓ & \textbf{No. Paths}↓ & \textbf{layer-wise}  \\ \midrule
& Vecfusion \cite{vectorfusion}         &31.10                   & 4668                  & 128.00        & False               \\ 
& SVGDreamer \cite{svgdreamer}        & 32.68                   & 5715                   & 512.00       & False              \\ 
& DiffSketcher \cite{diffsketcher}   & 31.47                   & 3374                   & 512.00       & False              \\ 
& Ours               & \textbf{33.76}                   & \textbf{27}                   & \textbf{35.39}         & True            \\ \midrule
\end{tabular}
\label{tab1}
\end{table}

\begin{table}[ht]
\centering
\footnotesize  
\caption{Comparison with SOTA Vectorization Methods.  The best results are denoted as \textbf{Bold}.}
\vspace{-3mm} 
\setlength{\tabcolsep}{4pt}
\begin{tabular}{lp{1.45cm}p{1.7cm}p{1cm}p{1cm}p{1cm}}
\toprule
& \textbf{Methods}   & \textbf{MSE}↓ & \textbf{Time Cost(s)}↓ & \textbf{No. Paths}↓ & \textbf{Layer-wise} \\ \midrule
& Diffvg \cite{diffvg}    &  $2.02 \times 10^{-4}$                  & 393                   & 256.00       & False               \\ 
& LIVE \cite{live}     & $5.21 \times 10^{-4}$                  & 3147                  & 46.00         & True             \\ 
& O\&R \cite{oar}     & $2.01 \times 10^{-4}$                  & 612                   & 64.00          & True            \\ 
& Ours      & \textbf{$1.96 \times 10^{-4}$ }        & \textbf{34}          & \textbf{29.98}    & True         \\ \bottomrule
\end{tabular}
\label{tab2}
\end{table}

\noindent \textbf{Quantitative Evaluation.}
Table \ref{tab1} and \ref{tab2} present the quantitative evaluation results. In the SVG generation task, our method achieves the highest CLIP-Score with the lowest average number of paths and shortest time cost. Notably, baseline methods fail to produce rationally layered outputs. For the layer-wise vectorization task, our approach outperforms all baselines across metrics: the lowest average path count demonstrates superior simplicity and efficiency, while faster runtime and higher reconstruction consistency further validate its effectiveness.


\begin{figure}[ht]
    \centering
    \includegraphics[width=1.0\linewidth]{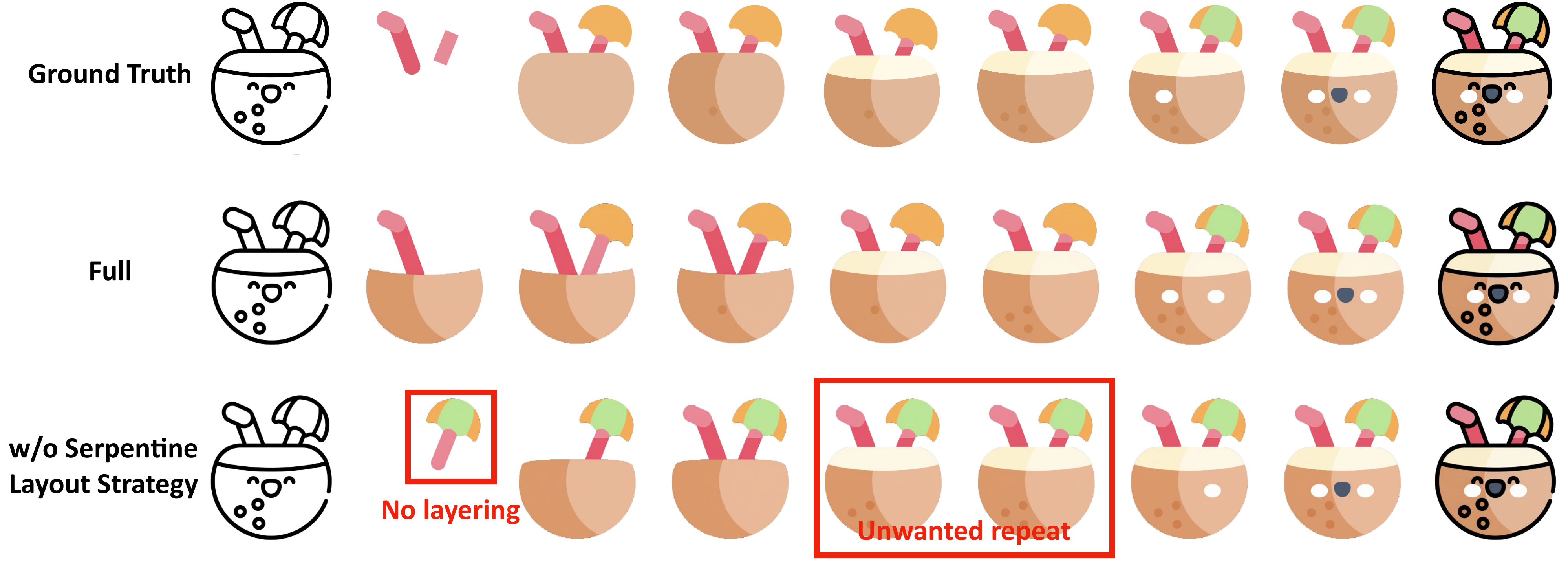} 
    \caption{Ablation study of Serpentine Layout Strategy.}
    \label{fig7}
\end{figure}

\begin{table}[ht]
\centering
\footnotesize  
\caption{Quantitative Evaluation of Serpentine Layout Strategy.}
\vspace{-3 mm}
\setlength{\tabcolsep}{2pt}
\begin{tabular}{lp{1.6cm}p{1.4cm}p{1.6cm}p{1cm}}
\toprule
\textbf{Methods} & \textbf{MSE\_train}↓ & \textbf{SSIM\_train}↑ & \textbf{MSE\_test}↓ & \textbf{SSIM\_test}↑ \\ \midrule
w/o Serpentine Layout         & $2.03 \times 10^{-4}$  & 0.964              & $2.41 \times 10^{-4}$ & 0.959 \\
Full                          & $1.65 \times 10^{-4}$  & 0.971              & $1.96 \times 10^{-4}$ & 0.963 \\
\bottomrule
\end{tabular}
\label{tab3}
\end{table}

\subsection{Ablation Study of Serpentine Layout Strategy.}
In this section, we conduct an ablation study on the serpentine layout strategy. As shown in Fig.~\ref{fig7}, when the serpentine layout strategy is not used to construct the training dataset, incomplete decomposition, undesirable repetitions, and abrupt changes between frames are more likely to occur. The quantitative evaluation results are presented in Table \ref{tab3}. For the layer-wise vectorization task, we calculate the MSE between the predicted results for 9 frames and the ground truth on both the training and test sets. When the serpentine layout strategy is not applied, the MSE is higher. This indicates that the Serpentine Layout Strategy benefits the model in learning a consistent layering process.





\begin{figure}[ht]
    \centering
    \includegraphics[width=0.78\linewidth]{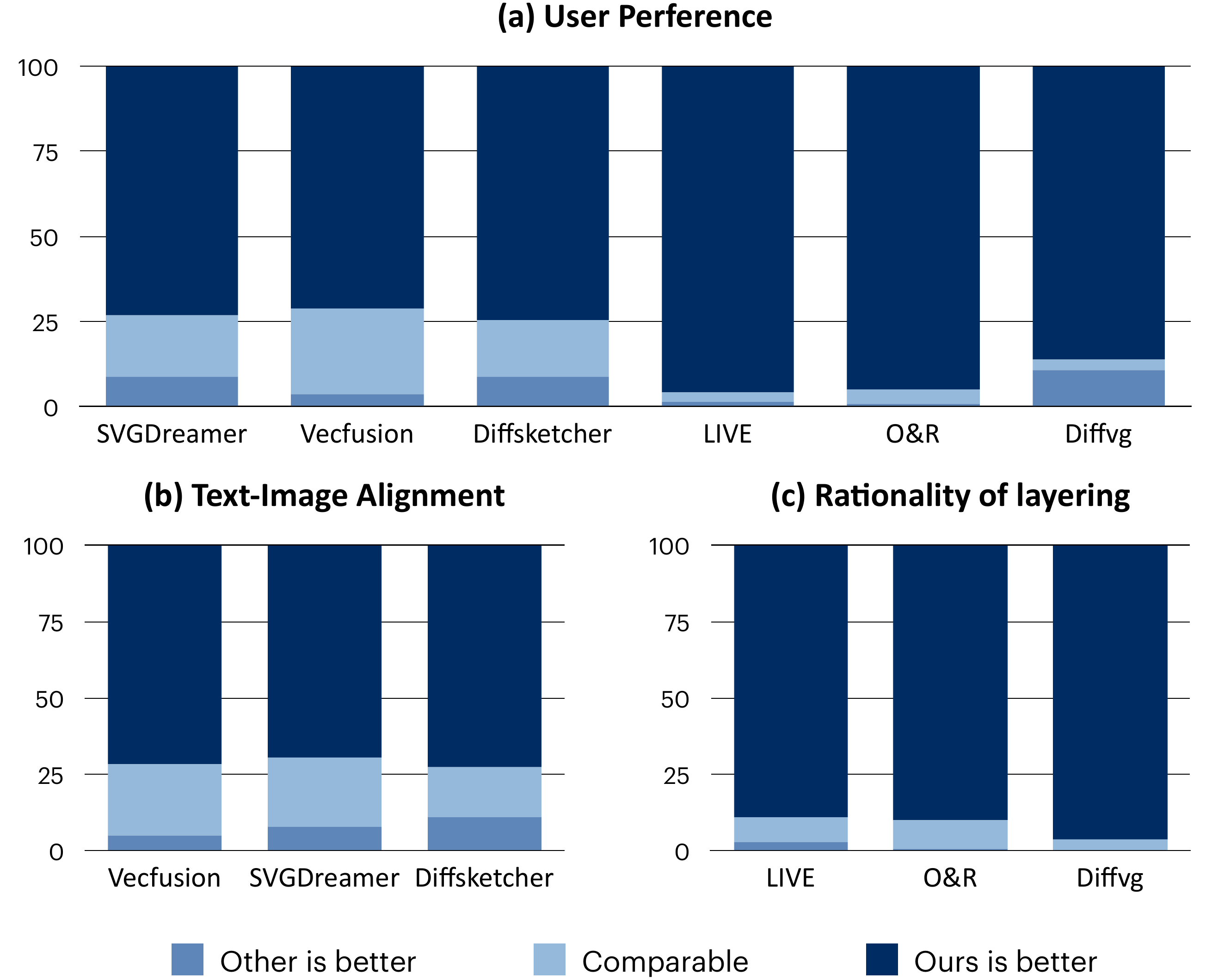} 
    \caption{User Study Results. LayerTracer outperform in all three metrics.}
    \label{fig8}
\end{figure}

\subsection{User Study}

We conducted a user study with 46 design enthusiasts using a digital questionnaire, presenting results from both our method and baseline methods. Participants selected their preferred results and those best matching the prompt descriptions in the text-to-SVG task. In the layer-wise vectorization task, they chose their preferred results and the most logically layered sequences. As Fig. \ref{fig8} shows, LayerTracer outperformed all baselines in user preference, prompt adherence, and layer rationality.

\section{Limitations and Future Work}
During layer-wise vectorization, LayerTracer relies on methods like Vtracer, inheriting its limitations such as  need for manual adjustment of hyperparameters. Its performance also falters on out-of-distribution data and complex images. We aim to develop a smarter single-layer vectorization solution to replace Vtracer in the future.



\section{Conclusion}

In this work, we introduced LayerTracer, a novel framework that bridges the gap between automated SVG generation and professional design standards. Leveraging the strengths of Diffusion Transformers, LayerTracer achieves  cognitive-aligned, layer-wise SVG generation and vectorization. By learning the workflows and design logic of human designers, LayerTracer effectively generates clean, editable, and semantically meaningful vector graphics from textual descriptions or raster images. To overcome the scarcity of layered SVG creation data, we established a pipeline that collects over 20,000 SVG creation sequences. We proposed Serpentin dataset construction method,  enabling effective model training. Extensive experiments demonstrate that LayerTracer not only excels in SVG generation quality but also offers unparalleled flexibility and interpretability, setting a new benchmark for scalable vector graphics creation.


\section*{Acknowledgement}
This research is supported by the National Research Foundation, Singapore under its AI Singapore Programme (AISG Award No: AISG3-RP-2022-030).

\newpage
\bibliographystyle{ACM-Reference-Format}
\bibliography{acmart}

\clearpage
\setcounter{page}{1}
\maketitlesupplementary
\renewcommand\thesection{\Alph{section}}
\setcounter{section}{0}  
\newcommand{\best}[1]{\textbf{#1}}

\section{User Study Details}
\label{Preliminary}

We conducted a user study in the form of an online survey, with a total of 46 participants evaluating 36 questions per questionnaire. The study was divided into two sections:

The first section evaluated text-to-vector generation results, comparing different methods. Each comparison included the corresponding text prompt, allowing participants to select the generated vector graphic that best reflected the original textual description. The evaluation consisted of three examples per method, leading to a total of nine comparisons, with two questions per comparison:

1. Which vector graphics result looks better?  
2. Which result better reflects the meaning of the original text?  

The second section focused on layer-wise vectorization, comparing different approaches. To ensure a fair evaluation, we provided a structured breakdown of the vectorization process, allowing participants to assess the quality and rationality of the vector graphic creation process. This section also contained three examples per method, resulting in nine comparisons, with two evaluation questions:

1. Which vectorization result looks better?  
2. Which result better preserves the characteristics of the original raster image?



\section{Dataset Construction Methods.}

The SVG dataset proposed in this work is collected from multiple sources, including the internet, vendor procurement, and designer-created assets. In addition to manually crafted layered SVGs, we also engaged designers to reorganize and arrange layers of other SVGs into a logical sequence that aligns with the design creation process. The final training data was obtained through multiple rounds of human-in-the-loop filtering. The dataset consists of a total of 20K samples across three styles: black outline icons (3K), illustrations (2K), and emojis (15K). For caption annotation, we added different triggers for different datasets and used the Florence-2 model \cite{Xiao2023Florence2AA} to label the last frame of the SVG sequence.

\begin{figure}[!t]
    \centering
    \includegraphics[width=0.9\linewidth]{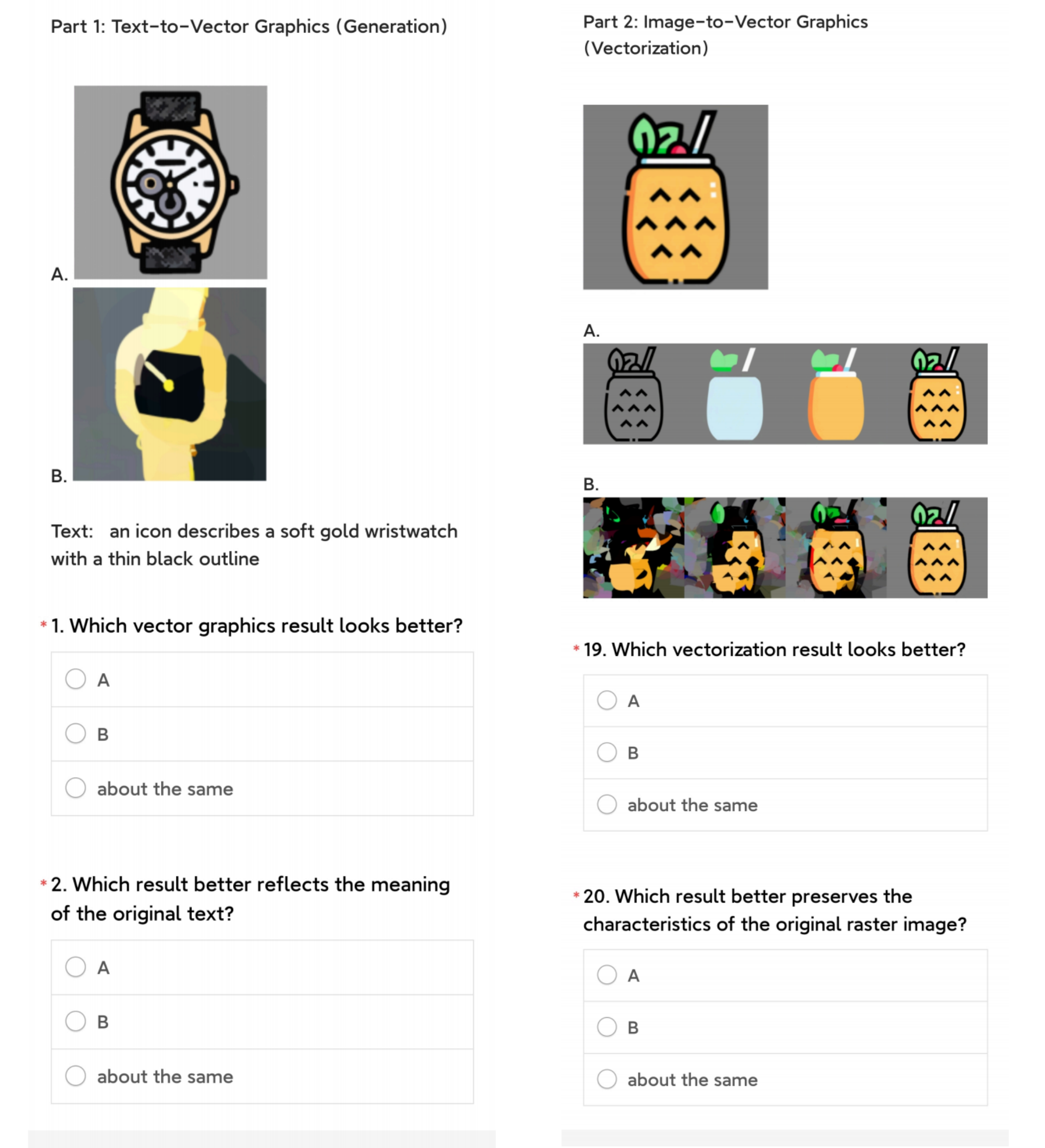}
    \caption{Examples of questions in the User Study online questionnaire.}
    \label{fig:data}
\end{figure}

\section{Failure Cases}

This section presents some failure cases of LayerTracer. As shown in the figure, when generating pixel-format sequence data, issues such as unwanted repeat, no layering, and inconsistencies between consecutive frames may occur. The problem of undesired repetition can be mitigated during the Layer-wise vectorization stage by comparing adjacent frames to remove duplicates. However, incorrect layering and frame inconsistencies can negatively impact the quality of Layer-wise SVG generation.

\begin{figure*}[!t]
    \centering
    \includegraphics[width=1.\linewidth]{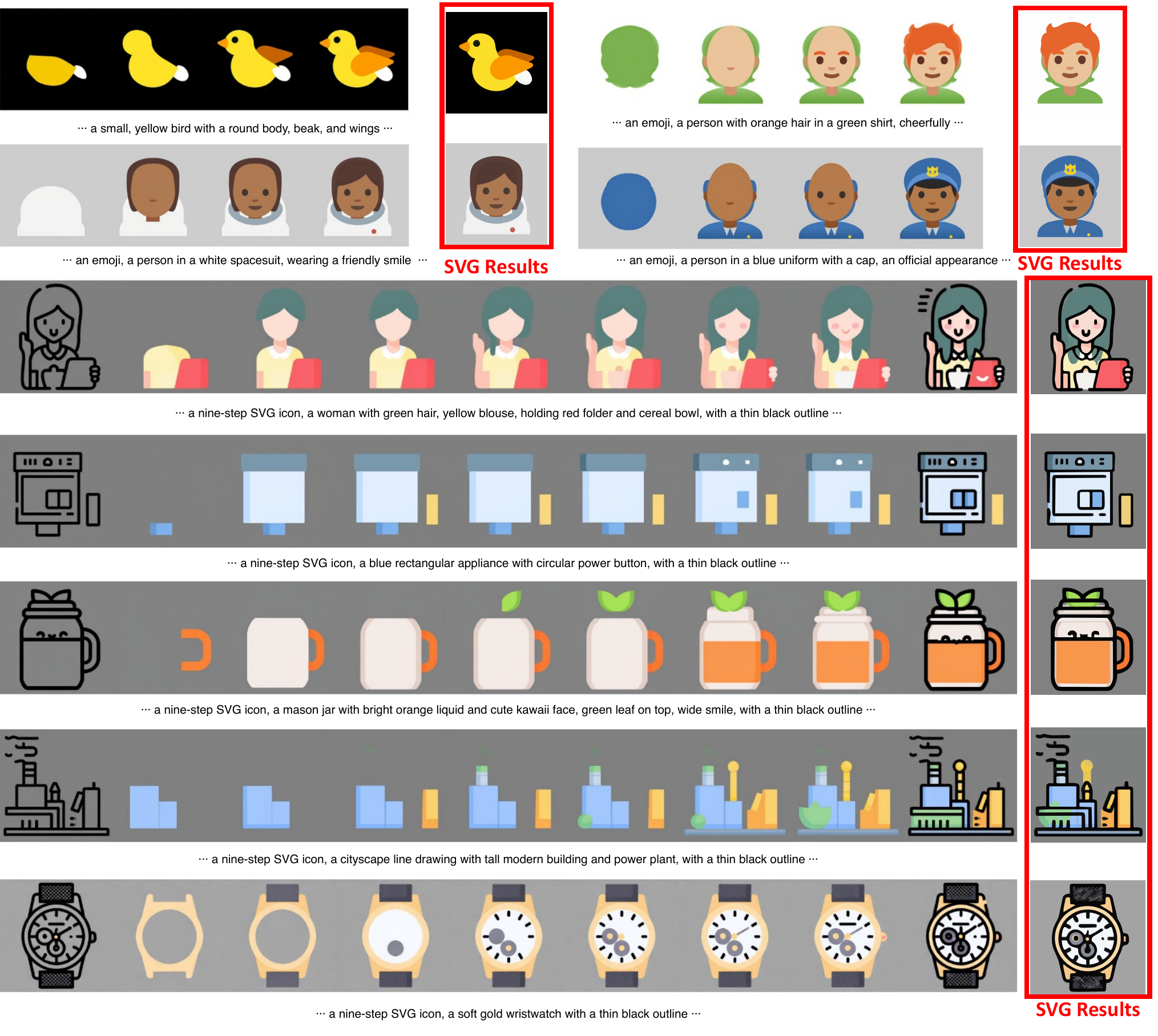}
    \caption{More results of layered vector graphics generation. The red boxes highlight the SVG format results, which can be zoomed in to view the details.}
    \label{fig:data}
\end{figure*}

\begin{figure*}[!t]
    \centering
    \includegraphics[width=1.\linewidth]{image/dataset.pdf}
    \caption{Each row represents examples of SVG datasets for three categories: emojis, illustrations, and black outline icons.}
    \label{fig:data}
\end{figure*}

\begin{figure*}[!t]
    \centering
    \includegraphics[width=1.\linewidth]{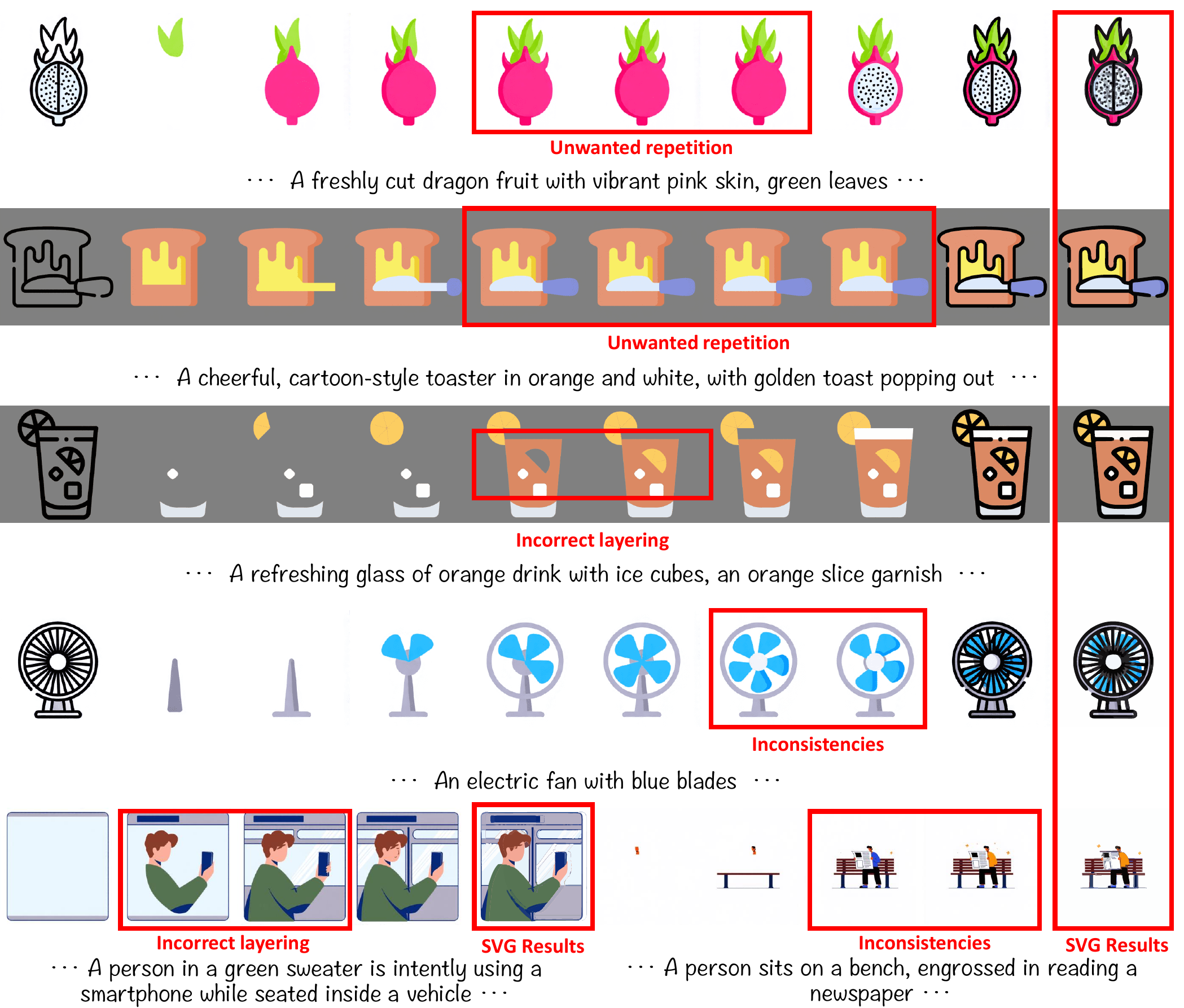}
    \caption{Some failure cases.}
    \label{fig:data}
\end{figure*}

\end{document}